\documentclass{article}
\usepackage[utf8]{inputenc}
\usepackage{mlspconf}
\usepackage{amsmath,graphicx}
\usepackage{nicefrac}
\usepackage{subcaption}
\usepackage{soul}
\usepackage{color}
\usepackage{hyperref}

\definecolor{darkgreen}{rgb}{0,0.6,0.2}

\title{Remote sensing image regression for heterogeneous change detection}

\name{Luigi T. Luppino$^{1}$, Filippo M. Bianchi$^{1}$, Gabriele Moser$^{2}$, Stian N. Anfinsen$^{1}$}
\address{Machine Learning Group, Department of Physics and Technology, UiT The Arctic University of Norway$^{1}$ \\ DITEN Department, University of Genoa, Italy$^{2}$ \\ \texttt{luigi.t.luppino@uit.no}}
\begin{document}

\ninept

\maketitle

\begin{abstract}
Change detection in heterogeneous multitemporal satellite images is an emerging topic in remote sensing.
In this paper we propose a framework, based on image regression, to perform change detection in heterogeneous multitemporal satellite images, which has become a main topic in remote sensing.
Our method learns a transformation to map the first image to the domain of the other image, and vice versa.
Four regression methods are selected to carry out the transformation: Gaussian processes, support vector machines, random forests, and a recently proposed kernel regression method called homogeneous pixel transformation.
To evaluate not only potentials and limitations of our framework, but also the pros and cons of each regression method, we perform experiments on two data sets.
The results indicates that random forests achieve good performance, are fast and robust to hyperparameters, whereas the homogeneous pixel transformation method can achieve better accuracy at the cost of a higher complexity.

\end{abstract}

%%%%%%%%%%%%%%%%%%%%%%%%%%%%%%%%%%%%%%%%%%%%%%%%%%%%%%%%%%%%%%%%%
%%%%%%%%%%%%%%%%%%%%%%%%% INTRODUCTION %%%%%%%%%%%%%%%%%%%%%%%%%%
%%%%%%%%%%%%%%%%%%%%%%%%%%%%%%%%%%%%%%%%%%%%%%%%%%%%%%%%%%%%%%%%%

\section{Introduction}

Change detection (CD) is a well known task in satellite remote sensing: the goal is to recognise changes in imagery acquired on the same location but at different times.
The applications range from disaster assessment to long term trend monitoring \cite{zhao2017discriminative,zhang2016cd,gong2016coupled}.
Most of the past works on CD assume that the satellite images are homogeneous, i.e.\ the data were collected by the same kind of sensors and using the same configurations and modalities \cite{zhao2017discriminative,zhang2016cd,gong2016coupled}.
\iffalse Slight differences (e.g.\ in lightning conditions) might have been considered, but only in cases for which it was possible to compensate for them \cite{tuia2013graph}. \fi
Even though there are techniques which mitigate the issues due to misalignments \cite{zhang2016cd,marcos2016geospatial,liu2016unsupervised}, co-registration is another fundamental assumption for CD: every pixel of the image at time one and its corresponding pixel of the image at time two are assumed to represent the exact same location on the earth.

The development of new sensors and the improvement of their capabilities has eventually brought the remote sensing community to consider the use of satellite images acquired under heterogeneous conditions \cite{zhao2017discriminative,gong2016coupled,liu2017change,luppino2017clustering}.
This has led to methods based on heterogeneous sources of data \cite{zhao2017discriminative,zhang2016cd,liu2017change,mercier2008conditional,prendes2015new,liu2018change}, also referred to as multi-source \cite{gong2016coupled, tuia2016multi}, multi-modal \cite{marcos2016geospatial}, multi-sensor or cross-sensor \cite{liu2016unsupervised,storvik2009combination,volpi2015spectral,touati2018energy} and information unbalanced data \cite{su2017deep}.
As already reviewed in \cite{gong2016coupled}, there is not a unique way to group CD methods.
However, the distinction between techniques aimed at homogeneous and heterogeneous data is clear.
In the latter case, the assumptions that the same physical quantities are measured, classes have always the same signatures, and data follow the same statistical behaviour are no longer valid \cite{luppino2017clustering}.
Without any additional steps, traditional homogeneous CD techniques cannot handle this \cite{luppino2017clustering,liu2018change}.
To overcome the problem, a possible preliminary step is either project data from both times into a common domain \cite{gong2016coupled,tuia2016multi,storvik2009combination,volpi2015spectral} or transfer data from the time one to the time two domain \cite{mercier2008conditional,liu2018change}.
These methodologies are related to topics such as domain adaptation, data transformation and transfer learning \cite{marcos2016geospatial,liu2017change,liu2018change,tuia2016multi,khan2017forest}.
Post-classification comparison represents an exception: the best classifier for the pre-event data and the best one for the post-event data are selected, then the classification maps are compared to find the pixels which do not belong to the same class at both times.
Clearly, the performance with this approach depends highly on the choice and the design of the two classifiers, as well as on the quality and the size of the training set \cite{zhao2017discriminative}.
The exponential increase of interest in deep learning has also lead to the development of novel methods based on deep learning architectures, both in the homogeneous \cite{khan2017forest,lyu2016learning} and in the heterogeneous case \cite{zhao2017discriminative,zhang2016cd,su2017deep}.
Most of these methods are examples of feature learning, since they exploit the capability of e.g. convolutional neural networks (and especially stacked denoising autoencoders) to infer spatial information from the data and consequently to learn a new representation of it.

In this work, we suggest a simple, yet effective methodology to perform CD with heterogeneously acquired data.
It consists of training a regression function to predict how every pixel at time one would have been if it was acquired by a second sensor at time two, and vice versa.
We will refer to this methodology as \textit{image regression}, a term which has on some occasions been used in the CD literature \cite{lunetta1999remote,singh1989review,mas1999monitoring}.
Once the predictions of the images are computed, homogenenous CD methods can be applied to obtain the map of changes.
In particular, we consider three supervised methods to perform the regression: Gaussian processes, support vector machines, and random forests.
Moreover, the homogeneous pixel transformation method recently proposed by Liu et al.\ \cite{liu2018change} is chosen as a representative of the state-of-the-art.
As main contribution, we evaluate the performance of the different regression methods in the proposed framework.
By testing a selection of both well-established and more recent regression methods on two different data sets, we evaluate the consistency of their performance, as well as their pros and cons, helping the user to choose the most suitable approach according to requirements, such as best performance in detection, easiest tuning of the hyperparameters, shortest training and test time.
The remainder of this article is the following: Section \ref{sec:method} introduces the reader to the methodology, the notation, and the regression methods listed above.
Results on two data sets are presented in Section \ref{sec:results}.
Section \ref{sec:concl} concludes the paper.

%Although it is designed to work for any possible couple of sensors, the discussion will focus on the case in which Synthetic Aperture Radar (SAR) and optical data are involved, since they are the most common and mainly used in this field of research \cite{zhao2017discriminative}.

%%%%%%%%%%%%%%%%%%%%%%%%%%%%%%%%%%%%%%%%%%%%%%%%%%%%%%%%%%%%%%%%%
%%%%%%%%%%%%%%%%%%%%%%%%%% METHODOLOGY %%%%%%%%%%%%%%%%%%%%%%%%%%
%%%%%%%%%%%%%%%%%%%%%%%%%%%%%%%%%%%%%%%%%%%%%%%%%%%%%%%%%%%%%%%%%

\section{Methodology}\label{sec:method}

We follow the notation adopted in \cite{liu2018change}: the two images represent the same region but are acquired by different sensors at a different times and are denoted as $\boldsymbol{X}$ and $\boldsymbol{Y}$, respectively.
A limited part of the image has changed between time one and time two.
A training data set $\mathcal{T}$ of $M$ corresponding pixel pairs, $\mathcal{T}=\{\left(\boldsymbol{x}_m,\boldsymbol{y}_m\right)\}_{m=1}^M$, is manually selected from areas not affected by changes in the two images.
According to \cite{liu2018change}, this provision of training data is not a strong requirement, although it prompts user interaction. 
The training data $\mathcal{T}$ allows us to learn a regression function $f^{(1)}$ such that
 \begin{equation}
 \boldsymbol{y}_m = f^{(1)}\left(\boldsymbol{x}_m\right) = \boldsymbol{\hat{y}}_m + \epsilon^{(1)}_m, \quad \ m=1,\dots,M
 \end{equation}
where $\boldsymbol{\hat{y}}_m$ is the dependent variable, $\boldsymbol{x}_m$ is the regressor, and $\epsilon^{(1)}_m$ is the residual.
We then train the reverse regression equation
 \begin{equation}
 \boldsymbol{x}_m = f^{(2)}\left(\boldsymbol{y}_m\right) = \boldsymbol{\hat{x}}_m + \epsilon^{(2)}_m, \quad \ m=1,\dots,M
 \end{equation}
in which $\boldsymbol{\hat{x}}_m$ is predicted starting from the regressor $\boldsymbol{y}_m$.
With these two functions it is possible to predict $\boldsymbol{\hat{Y}}$, i.e.\ the image which would have been obtained if sensor $\mathcal{Y}$ had observed the reality at time one, and $\boldsymbol{\hat{X}}$, the image of the reality at time two which would have been acquired by sensor $\mathcal{X}$.
Once the two predictions are computed, conventional change metrics such as image differences or ratios can be applied to highlight the differences between the original images and the corresponding predicted ones.
There is a plethora of more complex and more effective homogeneous CD techniques which could be applied at this stage, but the main goal of this work is to compare the image regression methods applied to obtain the predicted images.
The two-way regression can be referred to as an ensemble approach where two weaker results are combined to obtain a stronger and more reliable outcome.

\begin{figure}[h!]
\centering
\includegraphics[width=\columnwidth]{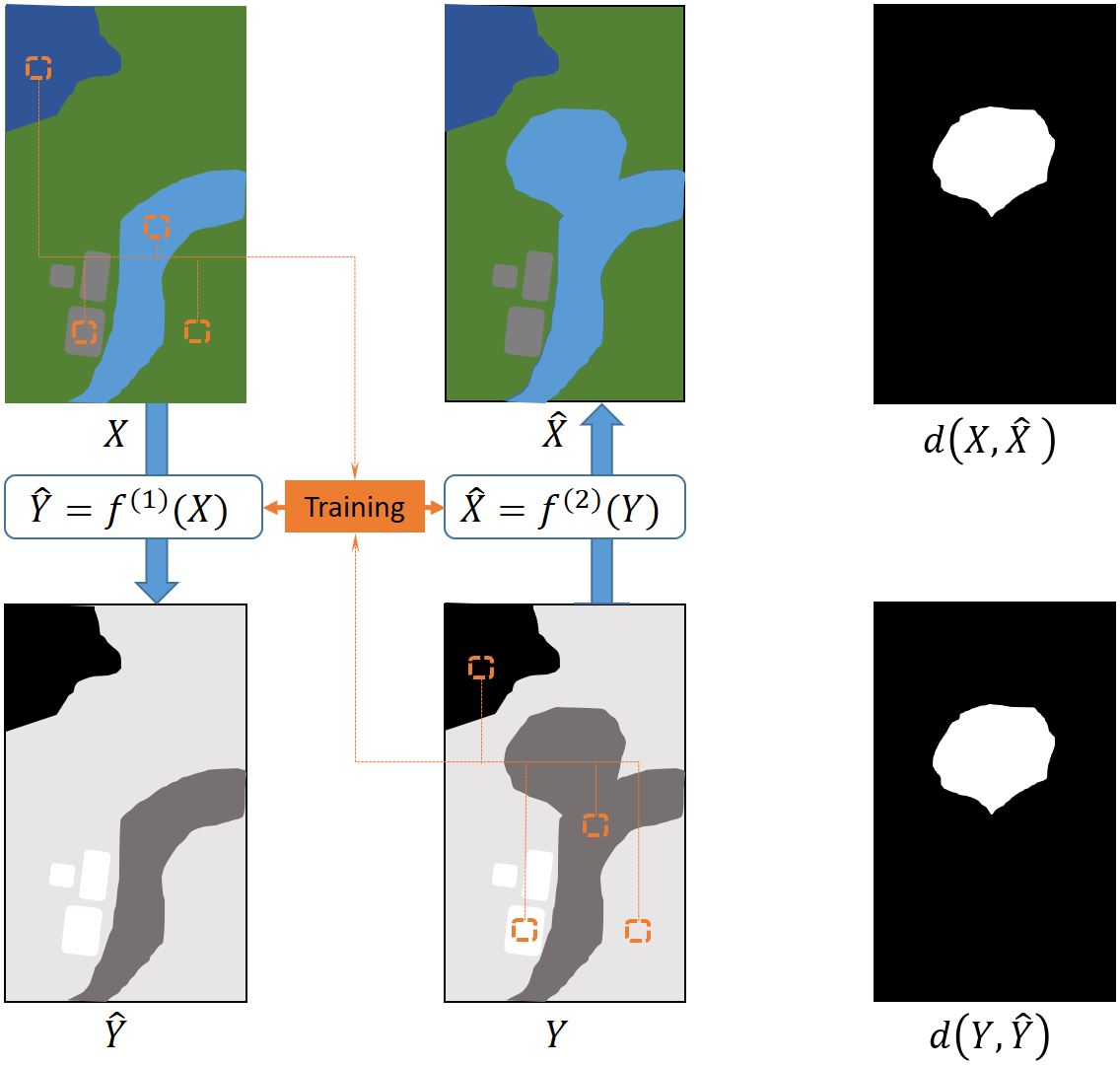}
\caption{Image regression: the two functions $f^{(1)}$ and $f^{(2)}$ are trained starting from the same data points, two predicted images are obtained, and finally two difference images are achieved.}
\label{fig:method}
\end{figure}

Let the distance image be defined as $$d(\cdot,\cdot):\mathbf{R}^{n1 \times n2 \times P} \times \mathbf{R}^{n1 \times n2 \times P} \longrightarrow \mathbf{R}^{n1 \times n2},$$ i.e.\ a pixel-wise distance between two images of size $n1\! \times\! n2$ and $P$ channels.
When the distance images, $d(\boldsymbol{X},\boldsymbol{\hat{X}})$ and $d(\boldsymbol{Y},\boldsymbol{\hat{Y}})$, are normalised and combined, distances that are consistently high in both images will indicate high probability of change, whereas false alarms due to a spurious high value in one of the distances will be suppressed.
We choose to combine the distances by a simple average. 
Before normalising, it is reasonable to clip the distances beyond some standard deviations of the mean value (e.g.\ $d_i > \bar{d}+4\sigma_d$), so that outliers do not compromise such a step.
At this stage, noise filtering can be applied if necessary.
Finally, a change map can be achieved by thresholding.
Fig.\ \ref{fig:method} illustrates the methodology.
In the following we briefly describe the regression methods considered in this work to evaluate $f^{(1)}$ and $f^{(2)}$.

%%%%%%%%%%%%%%%%%% GPR %%%%%%%%%%%%%%%%%%

\subsection{Gaussian Process Regression}

A Gaussian process (GP) is a collection of random variables, any finite subset of which have a joint Gaussian distribution.
It is completely specified by its mean function $\boldsymbol{m}\left(\boldsymbol{x}\right)$ and covariance (kernel) function $k_{\boldsymbol{x}_i,\boldsymbol{x}_j}=k\left(\boldsymbol{x}_i,\boldsymbol{x}_j\right)$.
For regression purposes, zero mean GPs are most often used \cite{rasmussen2004gaussian}.
Given a training set of $M$ input vectors (arranged in rows) $\boldsymbol{X}\! \in\! \mathbf{R}^{M\! \times\! P}$, the corresponding set of target vectors $\boldsymbol{Y}\! \in\! \mathbf{R}^{M\! \times\! Q}$, and a set of $N$ new observed vectors $\boldsymbol{X}_*\! \in\! \mathbf{R}^{N\! \times\! P}$, the joint distribution of the training vectors $\boldsymbol{Y}$ and the sought regressed vectors $\boldsymbol{\hat{Y}}\! \in\! \mathbf{R}^{N\! \times\! Q}$, conditioned on the input data $\boldsymbol{X}$ and $\boldsymbol{X}_*$, is
 \begin{equation}
\left[ \begin{tabular}{c}
    $\boldsymbol{Y}$ \\
    $\boldsymbol{\hat{Y}}$
\end{tabular}\right] \arrowvert\boldsymbol{X},\boldsymbol{X}_*
\sim \mathcal{N} \left(\boldsymbol{0},\left[\begin{tabular}{c c}
    $\boldsymbol{K}_{\boldsymbol{X},\boldsymbol{X}}$ &  $\boldsymbol{K}_{\boldsymbol{X},\boldsymbol{X_*}}$\\
    $\boldsymbol{K}_{\boldsymbol{X_*},\boldsymbol{X}}$ &  $\boldsymbol{K}_{\boldsymbol{X_*},\boldsymbol{X_*}}$
\end{tabular}\right]\right).
 \end{equation}
where $\boldsymbol{K}_{\boldsymbol{X},\boldsymbol{X_*}}$ is the matrix whose $(i,j)$th entry is the value of the covariance on the $i$th row of $\boldsymbol{X}$ and the $j$th row of $\boldsymbol{X}_*$, and $\boldsymbol{K}_{\boldsymbol{X},\boldsymbol{X}}$, $\boldsymbol{K}_{\boldsymbol{X_*},\boldsymbol{X}}$, and $\boldsymbol{K}_{\boldsymbol{X_*},\boldsymbol{X_*}}$ have similar meanings.
Thus, the following posterior distribution is derived (see \cite{rasmussen2004gaussian} for details):
 \begin{equation}
\begin{split}
\boldsymbol{\hat{Y}}\! \mid\! \boldsymbol{X_*},\boldsymbol{X},\boldsymbol{Y}\! \sim\! \mathcal{N}\! \big(\! & \boldsymbol{K}_{\boldsymbol{X_*},\boldsymbol{X}}\cdot\boldsymbol{K}_{\boldsymbol{X},\boldsymbol{X}}^{-1}\cdot\boldsymbol{Y}, \\ & \boldsymbol{K}_{\boldsymbol{X_*},\boldsymbol{X_*}}-\boldsymbol{K}_{\boldsymbol{X_*},\boldsymbol{X}}\cdot\boldsymbol{K}_{\boldsymbol{X},\boldsymbol{X}}^{-1}\cdot\boldsymbol{K}_{\boldsymbol{X},\boldsymbol{X_*}}\big)
\end{split}
 \end{equation}
The two main factors affecting the quality of the regression are the choice of kernel function and its hyperparameters.
In this work, we opted for the commonly used radial basis function (RBF)
 \begin{equation}
k_{\boldsymbol{x}_i,\boldsymbol{x}_j}=\sigma_f^2\exp\left(-\frac{1}{2}\left(\boldsymbol{x}_i-\boldsymbol{x}_j\right)^TL\left(\boldsymbol{x}_i-\boldsymbol{x}_j\right)\right)\,,
 \end{equation}
where $\boldsymbol{\theta} = \left\{L,\sigma_f^2\right\}$ is the set of hyperparameters, with signal variance $\sigma_f^2$ and $L = l^{-2}I$, if the length-scale parameter $l$ is a scalar (isotropic kernel), or $L = \text{diag}\left(\boldsymbol{l}^{-2}\right)$, if $\boldsymbol{l}$ is a vector (anisotropic kernel) \cite{rasmussen2004gaussian}.
Concerning the optimisation of $\boldsymbol{\theta}$, a gradient ascent is performed to maximise the marginal likelihood $\mathcal{P}\left(\boldsymbol{Y}\mid\boldsymbol{X},\boldsymbol{\theta}\right)$. %\commentL{I do not remember, is it this one? Sorry, I always do a mess with pdfs}
%
% \begin{equation}
%\mathcal{L}\left(\boldsymbol{Y}\mid\boldsymbol{X}\right)=\int_{\Omega_{\boldsymbol{\theta}}} \mathcal{P}\left(\boldsymbol{Y}\mid\boldsymbol{X},\boldsymbol{\theta}\right)\mathcal{P}\left(\boldsymbol{\theta}\right) d\boldsymbol{\theta}\ .
% \end{equation}
%
A weak point of this optimisation is that it might lead to a local maximum instead of the global one, so it is recommended to iterate the procedure several times starting from different random points in the hyperparameter space $\Omega_{\boldsymbol{\theta}}$.

%%%%%%%%%%%%%%%%%% SVM %%%%%%%%%%%%%%%%%%

\subsection{Multi-output Support Vector Regression}

Support vector machines (SVMs) are a very well known machine learning approach used for classification and regression.
By solving the so-called dual problem, it is possible to find the best separating or fitting curve with respect to a loss function that accounts for misclassification or reconstruction error and with respect to a regularisation parameter which defines the width of a soft margin around such a curve. In addition, the support vectors, i.e.\ the training points that define the 
margin, are highlighted from the rest of the training set.

Instead of coping with multi-output problems all at once, the solution usually adopted is to tune a different SVM for each regressand variable.
Therefore, the standard implementations of support vector regression (SVR) are designed to predict a single output feature, ignoring the potentially nonlinear relations across the target features \cite{tuia2011multioutput}.
Tuia et al.\ \cite{tuia2011multioutput} proposed a multi-input multi-output (MIMO) SVR method to overcome this limitation. During the training phase, it aims to minimise the cost function
 \begin{equation}
L_p\left(\boldsymbol{W},\boldsymbol{b}\right) = \frac{1}{2}\sum_{q=1}^Q\parallel\boldsymbol{w}_q\parallel^2+C\sum_{m=1}^M L\left(\mu_m\right)
 \end{equation}
where
\begin{align}
\label{eq:errorterms}
 L \left(\mu_m\right) & =
\begin{cases}
    0 & \mu_m < \epsilon \\
    \mu_m^2-2\mu_m\epsilon+\epsilon^2 & \mu_m \geq \epsilon
\end{cases} \ , \\
 \mu_m & = \ \parallel\boldsymbol{e}_m\parallel \ =\sqrt{\boldsymbol{e}_m^T\boldsymbol{e}_m} \ , \\
 \boldsymbol{e}_m^T & = \boldsymbol{y}_m^T\! -\phi\! \left(\boldsymbol{x}_m\right)^T\boldsymbol{W}\! -\boldsymbol{b}^T.
\end{align}
Here, $\boldsymbol{W}=\left[\boldsymbol{w}_1,\dots,\boldsymbol{w}_Q\right],\, \boldsymbol{w_q}\! \in\! \mathbf{R}^{P}$ and $\boldsymbol{b}=\left[q_1,\dots,q_Q\right]$ are the coefficients and the bias of the linear combination of the data points $\boldsymbol{x}_m$ transferred in the Hilbert space by the kernel function $\phi$.
The penalty factor $C$ sets the trade-off between the regularisation term and the sum of the error terms $L\left(\mu_m\right)$.
If it is too large, nonseparable points would highly penalise the cost function and too many data points would turn into support vectors, causing overfitting. Vice versa, a small $C$ would lead to underfitting.
$\epsilon$ is half the width of the insensitivity zone.
This zone delimits a "tube" around the approximated function and all the training data points within the insensitivity zone do not contribute to the cost function (see Eq.\ \ref{eq:errorterms}).
For too small values of $\epsilon$, too many data points would be considered as support vectors (overfitting), the generalisation performance would be affected and the variance of the fitted curve would be too large.
On the contrary, a too large $\epsilon$ would cause underfitting and the overall accuracy would be low.
Another critical hyperparameter is the width $\sigma$ of the RBF kernel $\phi$.
To select the right combination of hyperparameters $\boldsymbol{\theta} = \left\{C,\epsilon,\sigma\right\}$, a grid search for the smallest cross-validation error or the minimization of an error bound can be applied.

%%%%%%%%%%%%%%%%%% RF %%%%%%%%%%%%%%%%%%

\subsection{Random Forest Regression}

Random forests (RF) were proposed by Breiman in \cite{breiman2001random} to perform both classification and regression, by exploiting the simplicity of random decision trees and the robustness of ensemble methods. From now on, only regression will be considered, but for classification purposes the approach is similar.

A RF consists on $T$ trees, at whose nodes $m$ randomly selected features are compared to random thresholds (e.g.\ $\text{feat}_1 > \text{thr}_1 \ \& \dots \& \ \text{feat}_m > \text{thr}_m)$.
In each tree, the training data points are divided over the branches according to these conditions, and the trees expand until only one data point is contained in each of the final nodes (leaves).
Thus, the corresponding training vectors $\boldsymbol{y}_m$ are assigned to the leaves.
During the test phase an input vector $\boldsymbol{x_*}$ goes through each tree and reaches one of the leaves, giving as output the assigned training vector $\boldsymbol{y}_m$.
Finally, the average of the $T$ outputs is computed, thereby obtaining the final regressed vector $\boldsymbol{\hat{y}}$.

To generalise better, every tree is trained on a bootstrap sample drawn from the training set, and a randomly drawn subset of features (of fixed cardinality) is used on each node of each tree.
The validation is carried out through out-of-bag estimation \cite{breiman2001random}.
Moreover, the behaviour of a RF can be controlled by tuning three parameters: the size of the forest (i.e.\ the number of trees $T$), the number of features $m$ considered in every node, and the depth of the trees.
A common remedy against overfitting is to prune the trees by leaving $p > 1$ data points in each leaf node, which will give in output the average of their corresponding training vectors $\boldsymbol{y}_m$.
Concerning the number of features considered at every node, in \cite{breiman2001random} it is suggested by empirical results to set $m=\lfloor\frac{\log P}{\log 2}\rfloor$, where $P$ is the dimension of the vectors $\boldsymbol{y}$.
It is common practice to follow the rule of thumb: $m=\lfloor\nicefrac{P}{3}\rfloor$ \cite{breiman2001random}.
However, there are no practical rules to choose the size of the forest.
One may think that for a larger number of trees the outcomes become better, but \cite{breiman2001random} proved that at some point the overall accuracy saturates due to the rise of a strong correlation between the trees.
Therefore, a compromise between gained accuracy and computational load must be found.

%%%%%%%%%%%%%%%%%% HPT %%%%%%%%%%%%%%%%%%

\subsection{Homogeneous Pixel Transformation}

The homogeneous pixel transformation (HPT) method proposed by Liu et al.\ \cite{liu2018change} is a kernel regression based on the K-nearest neighbours (KNN) of each data point.
This technique recalls the distance weighted averaging or locally weighted regression previously presented in \cite{atkeson1997locally}, where many related aspects are also studied: possible kernels, distance measures, choices of the bandwidth, denoising techniques, and outlier detection.
For every data point in the first image $\boldsymbol{x}_i$, the $K$ nearest neighbours among the training vectors $\boldsymbol{x}_m \in \mathcal{T}$ are sought for. The regression consists of the weighted sum
 \begin{equation}
 \boldsymbol{\hat{y}}_i = \sum_{k=1}^{K}w_{i,k}\cdot \boldsymbol{y}_{i,k}\ ,
 \end{equation}
 where
  \begin{equation}
 w_{i,k} = w\left(\boldsymbol{x}_i,\boldsymbol{x}_k\right) = e^{-\gamma d_{i,k}}\ .
 \end{equation}
$d_{i,k}$ is the Euclidean distance between $\boldsymbol{x}_i$ and its $k^{th}$ nearest neighbour $\boldsymbol{x}_k$, $\boldsymbol{y}_{i,k}$ is the corresponding vector of $\boldsymbol{x}_k$ in $\mathcal{T}$, whereas the kernel width $\gamma$ regulates how strongly the farthest neighbours are penalised.
If $\gamma$ is too small, the addends tend to be equally weighted and the sum is close to an average, if $\gamma$ is too large, few main addends contribute to the sum whilst the rest are heavily penalised.
Before computing the weights, a relative normalisation of the distances is applied:
 \begin{equation}
 d_{i,k} = \frac{\left\Arrowvert\boldsymbol{x}_i-\boldsymbol{x}_k\right\Arrowvert}{\max_k\left\Arrowvert\boldsymbol{x}_i-\boldsymbol{x}_k\right\Arrowvert}\ .
 \end{equation}
The normalisation in \cite{liu2018change} is defined as relative, because it considers the maximum among the distances between the data point $\boldsymbol{x}_i$ and its neighbours.
However, while testing our implementation, we found that it is better to perform an absolute normalisation, thus seeking the maximum among all the computed distances.

%%%%%%%%%%%%%%%%%%%%%%%%%%%%%%%%%%%%%%%%%%%%%%%%%%%%%%%%%%%%%%%%%
%%%%%%%%%%%%%%%%%%%%%%%%%% EXPERIMENTS %%%%%%%%%%%%%%%%%%%%%%%%%%
%%%%%%%%%%%%%%%%%%%%%%%%%%%%%%%%%%%%%%%%%%%%%%%%%%%%%%%%%%%%%%%%%

\begin{figure*}[h!]

\begin{subfigure}[t]{0.19\textwidth}
\includegraphics[width=\linewidth,keepaspectratio,trim ={6.3cm 4.9cm 6.3cm 2.3cm}, clip]{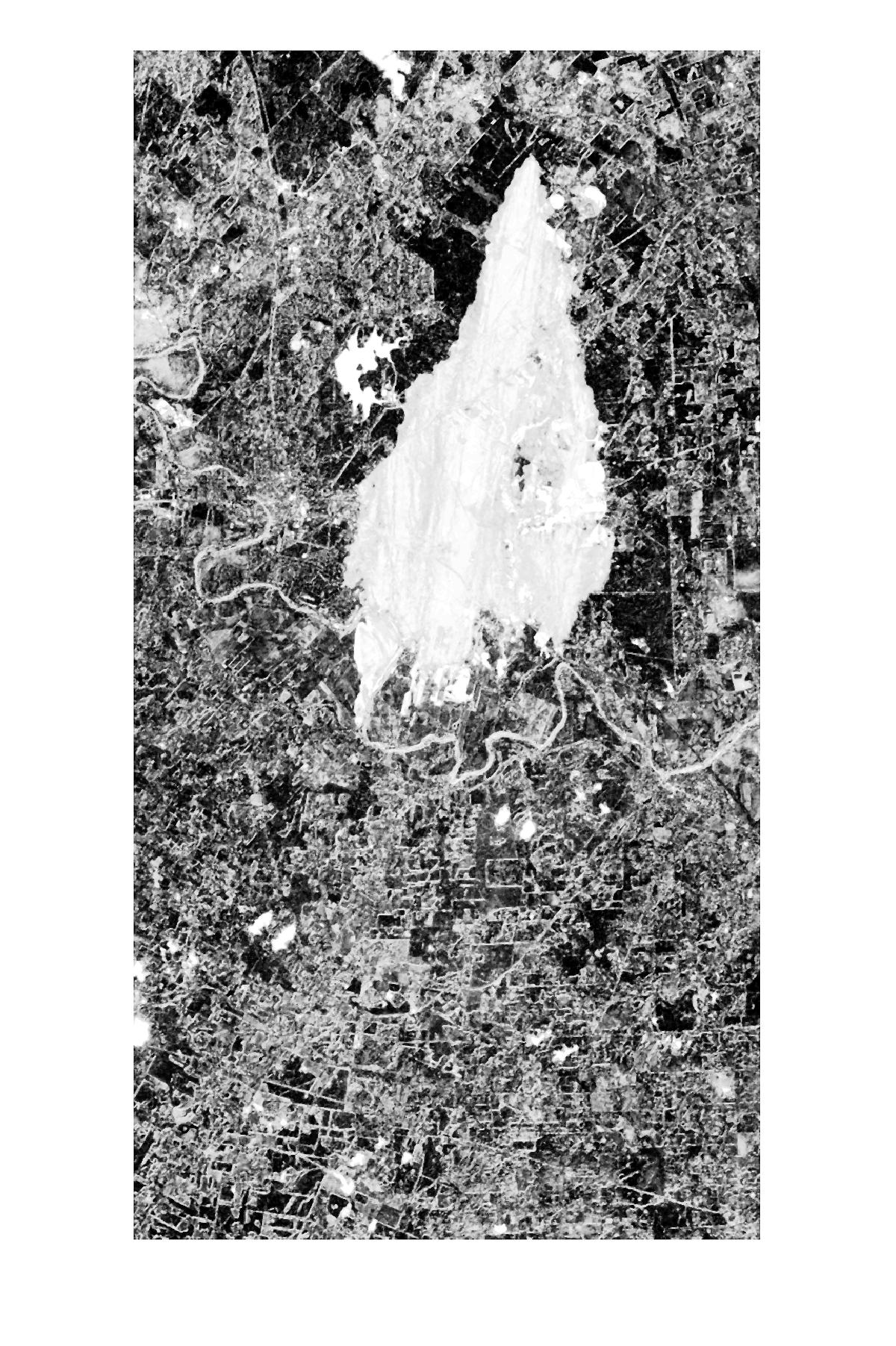} 
\caption{GP}
\label{fig:subim1}
\end{subfigure}
\begin{subfigure}[t]{0.19\textwidth}
\includegraphics[width=\linewidth,keepaspectratio, trim = {6.3cm 4.9cm 6.3cm 2.3cm}, clip]{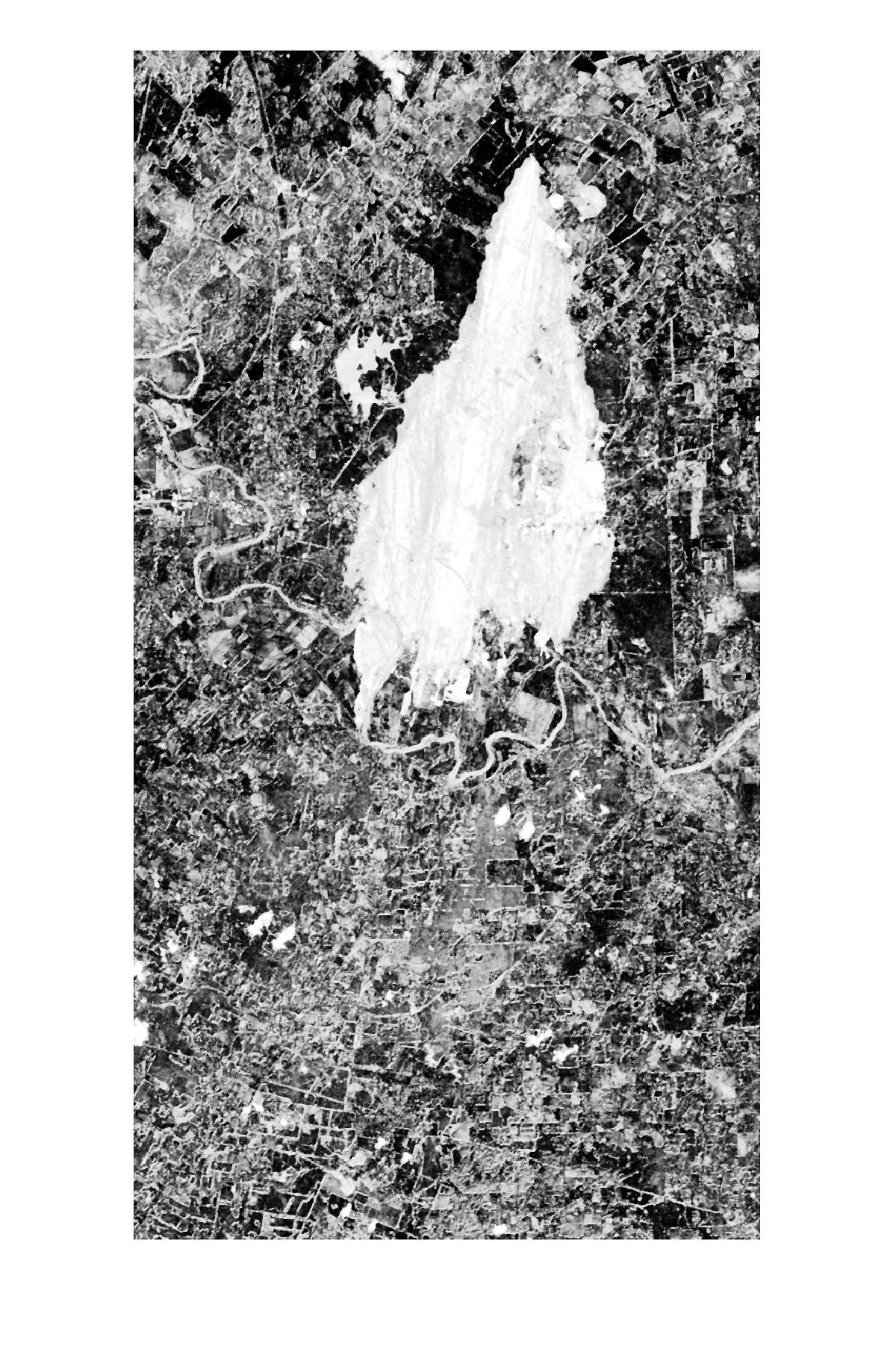}
\caption{SVR}
\label{fig:subim2}
\end{subfigure}
\begin{subfigure}[t]{0.19\textwidth}
\includegraphics[width=\linewidth,keepaspectratio, trim = {6.3cm 4.9cm 6.3cm 2.3cm}, clip]{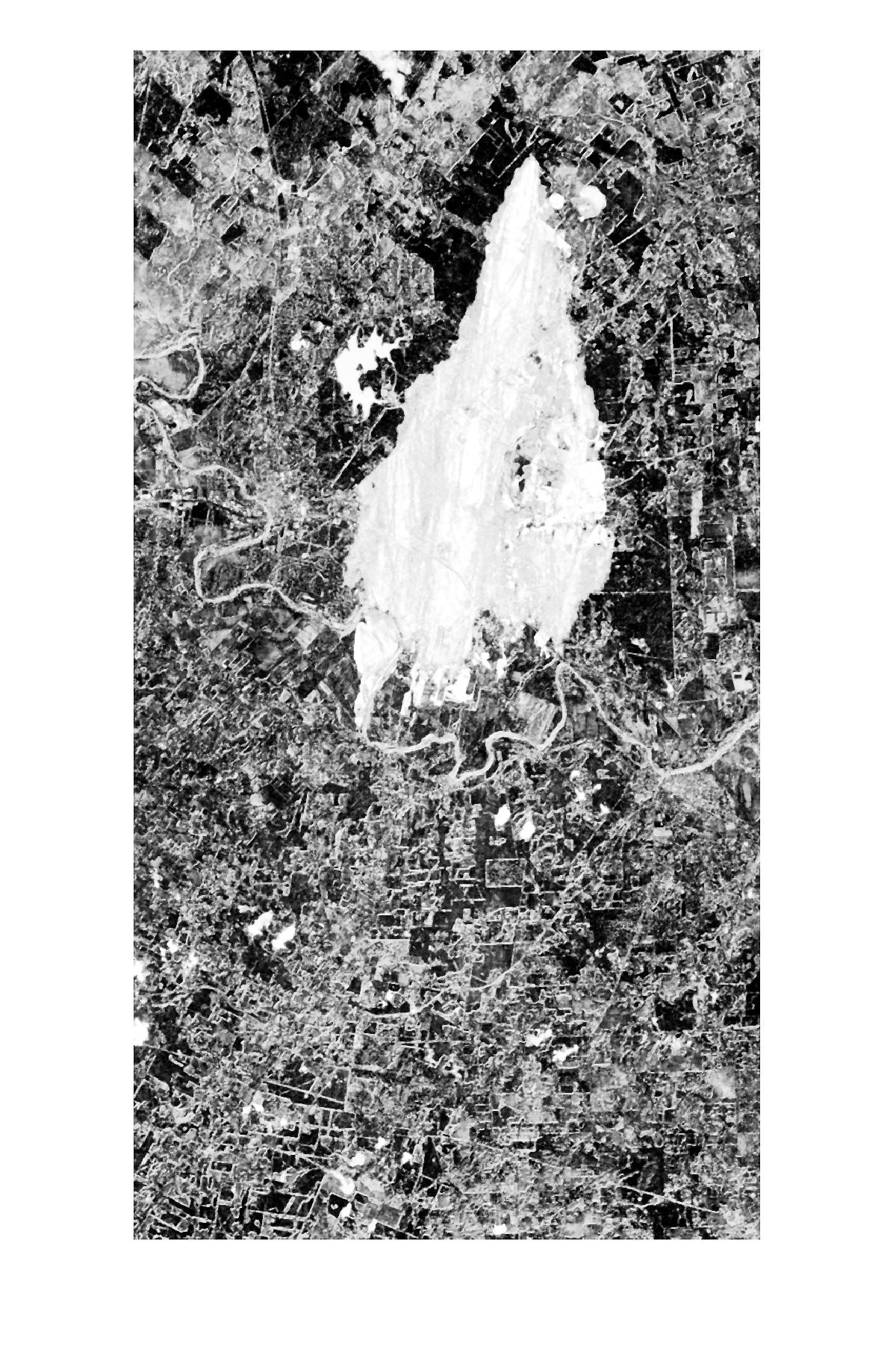}
\caption{RF}
\label{fig:subim3}
\end{subfigure}
\begin{subfigure}[t]{0.19\textwidth}
\includegraphics[width=\linewidth,keepaspectratio, trim = {6.3cm 4.9cm 6.3cm 2.3cm}, clip]{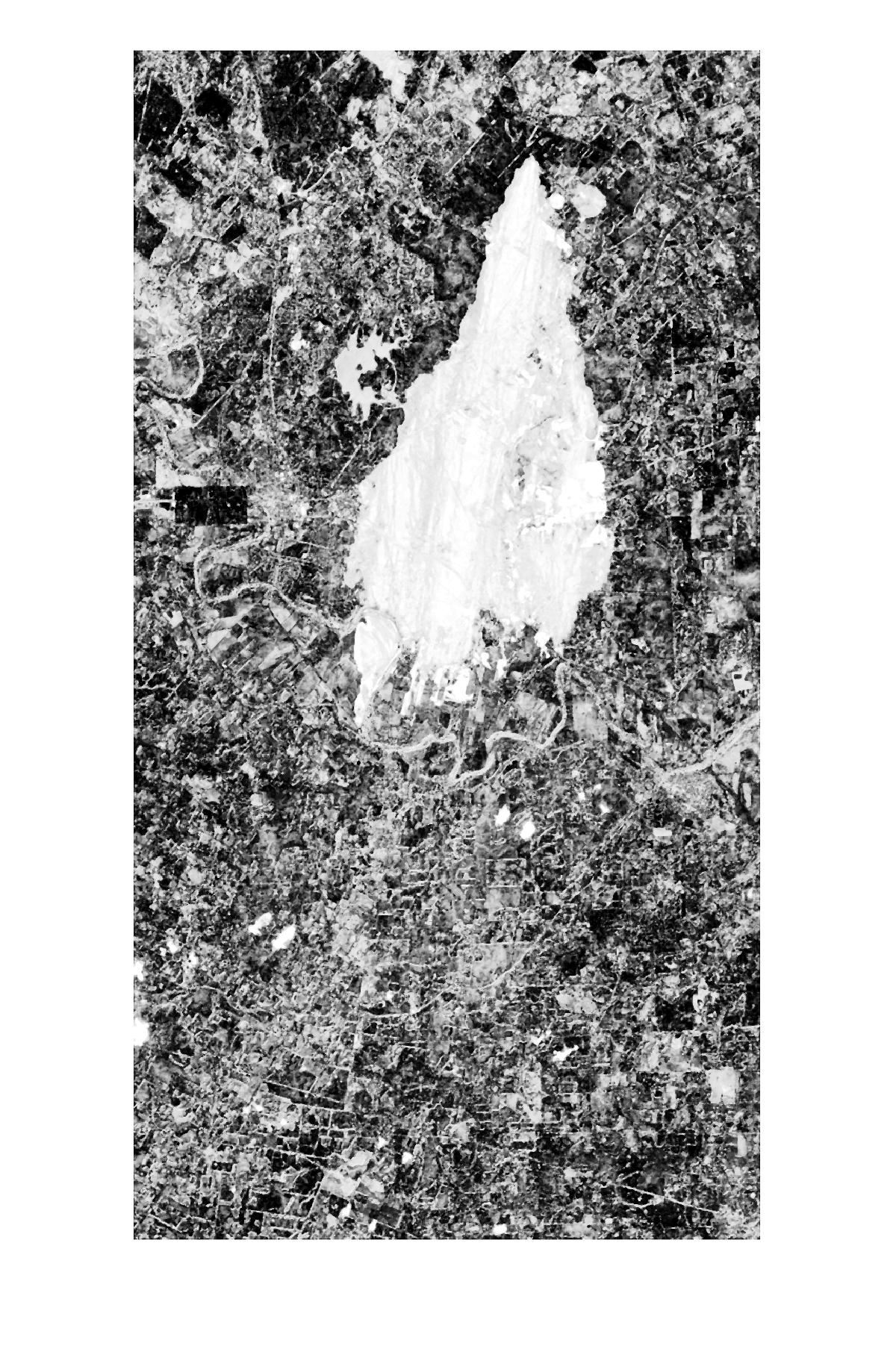}
\caption{HPT}
\label{fig:subim4}
\end{subfigure}
\begin{subfigure}[t]{0.19\textwidth}
\includegraphics[width=\linewidth,keepaspectratio, trim = {6.3cm 4.9cm 6.3cm 2.3cm}, clip]{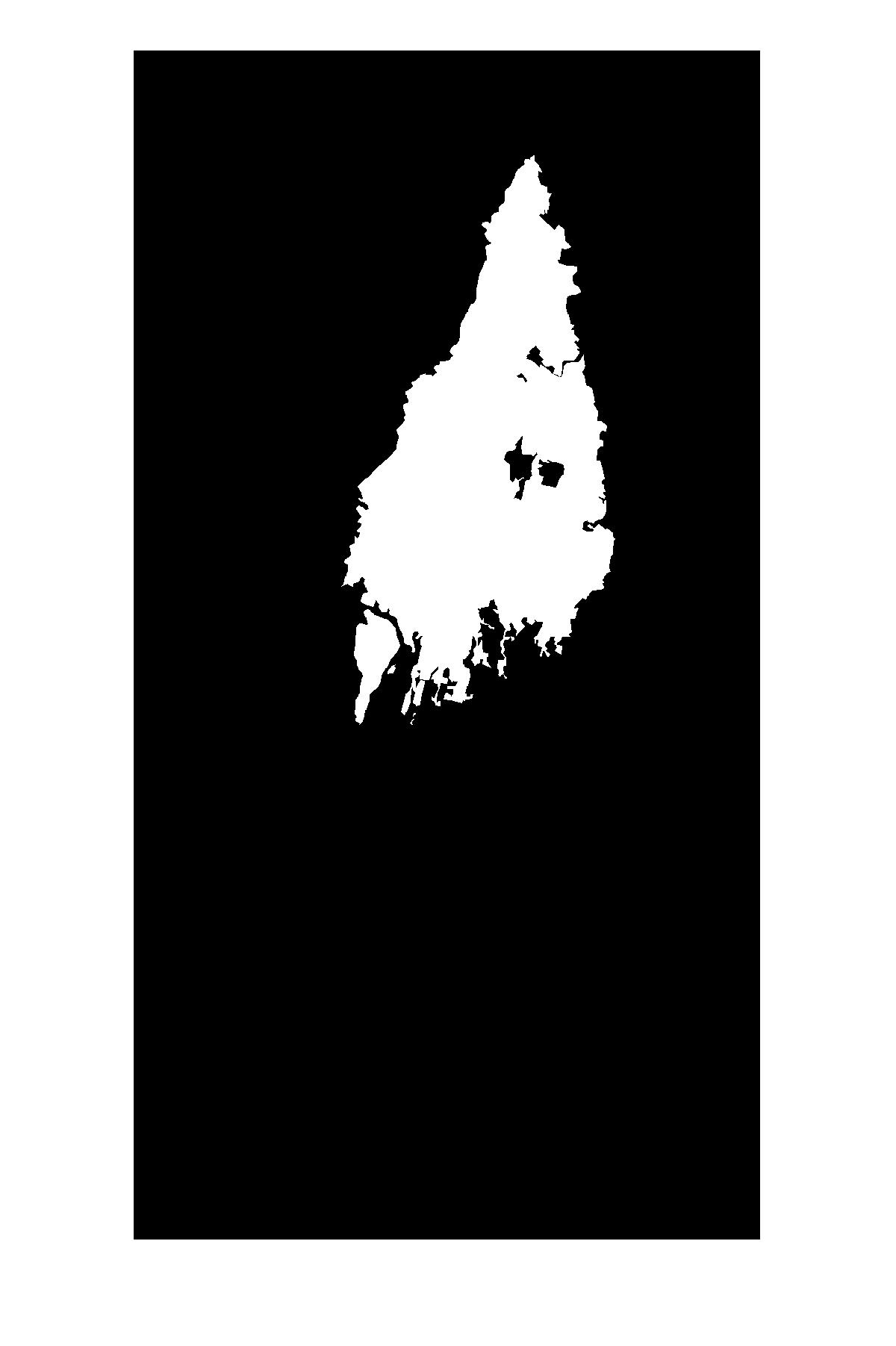}
\caption{Ground Truth}
\label{fig:subim5}
\end{subfigure}

\caption{Distance images obtained by applying the proposed approach with different regression methods: (a) Gaussian process regression, (b) MIMO support vector regression, (c) random forest regression, (d) homogeneous pixel transformation \cite{liu2018change}, (e) ground truth.}
\label{fig:dataset1}
\end{figure*}

\section{Experimental results}\label{sec:results}

The performance of the CD framework, configured with the proposed regression methods, is evaluated on two different data sets in terms of accuracy and computational speed.
Accuracy is measured in terms of \emph{area under the curve} (AUC), a value between $0.5$ (poor) and $1$ (optimal) which indicates the area below the receiver operating characteristic curve, which plots the false positive rate against the true positive rate.
The measured speed is the elapsed time during computation of the regressions in both the directions, starting from the training phase and ending after the test phase.
It must be pointed out that two of the methods are implemented in Python libraries (GP and RF), whereas the code provided by \cite{tuia2011multioutput} for the MIMO SVR method is written in MATLAB, and so is our implementation of the HPT method.
Therefore, an exact comparison of execution time of each algorithm is not possible, even though the two programming languages yield similar performance.
Nevertheless, the running times are indicators that can help us rank the four algorithms in terms of speed.

\subsection{Forest fire in Texas}

The first data set is composed of a multispectral image acquired by Landsat 5 TM before a forest fire in Bastrop County, Texas, during September-October, 2011.
An EO-1 ALI multispectral acquisition after the event completes the data set\footnote{Distributed by LP DAAC, \url{http://lpdaac.usgs.gov}}.
Both images are optical with $7$ and $10$ channels, respectively, some of which cover the same spectral bands, so the signatures of the classes involved are very similar.
Among the possible heterogeneous CD scenarios, this is one of the easiest.
The ground truth of the event (see Fig.\ \ref{fig:subim5}) is provided by Volpi et al.\ \cite{volpi2015spectral}.
The training set (roughly 2\% of the total data points) is selected manually with several rectangular patches taken from areas not affected by the fire event.
In a preliminary study phase, the hyperparameters of the GPs are set after only one iteration of the gradient ascent. 
For the SVR,  $C=1$, $\epsilon=0.1$, and $\sigma=1$ are set following \cite{tuia2011multioutput}.
Concerning the RF, $T = 128$, $m=[\nicefrac{P}{3}]$, $p = 5$ are chosen after a coarse grid search on $T$ and $p$.
Last, the HPT is tuned by setting $K = 300$ and $\gamma = 100$, as empirically found in \cite{liu2018change}.

After the combination of the two image differences, a $3 \times 3$ median filter is applied to remove salt and pepper noise.
In Fig.\ \ref{fig:dataset1}, the outcomes of the median filter for the four methods are depicted, showing how well they all behave.
The only exception is the HPT, which tends to overfit, as can be noticed in Fig.\ \ref{fig:subim4}.
One example is the black rectangular patch on the left of the area interested by the event, which actually corresponds to one of the selected patches of the training set.
However, all the AUCs reach values above 98\%, showing how well all four methods tackle the image regression task.
On one hand, this image pair is not especially challenging for the proposed approaches.
On the other hand, conventional CD methods designed for homogeneous data would be unfeasible here.
Hence, this result demonstrate the effectiveness of the proposed regression-based approach to heterogeneous CD.

\subsection{Flood in California}

\begin{figure*}[h!]

\begin{subfigure}[t]{0.24\textwidth}
\includegraphics[width=\linewidth,height=1.8\linewidth,trim ={6.3cm 4.9cm 6.3cm 2.3cm}, clip]{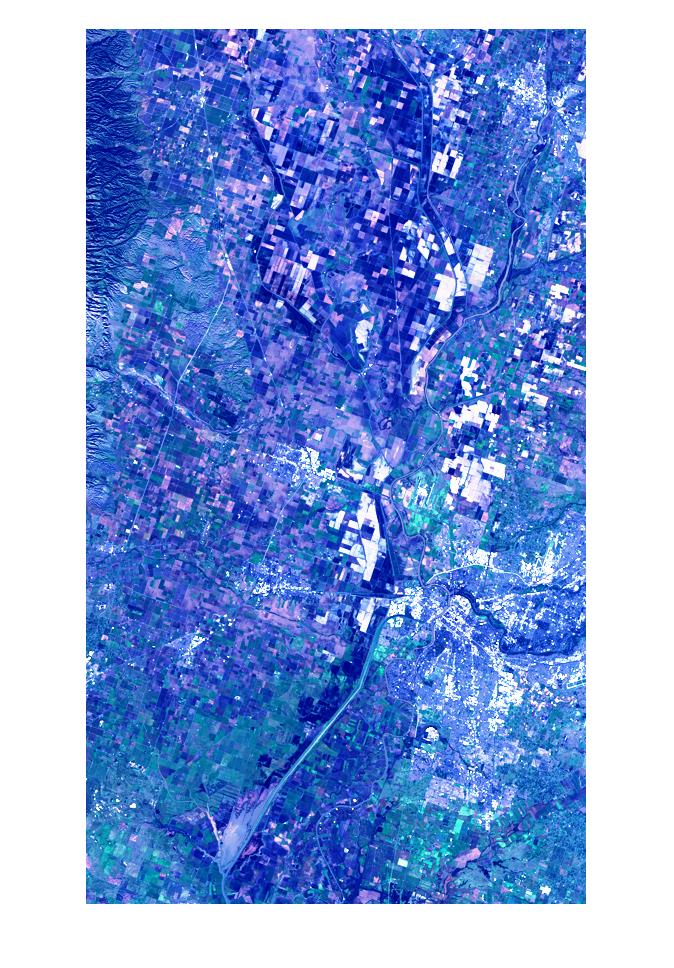} 
\caption{Landsat 8 ($t1$)}
\label{fig2:subim1}
\end{subfigure}
\begin{subfigure}[t]{0.24\textwidth}
\includegraphics[width=\linewidth,height=1.8\linewidth, trim = {6.3cm 4.9cm 6.3cm 2.3cm}, clip]{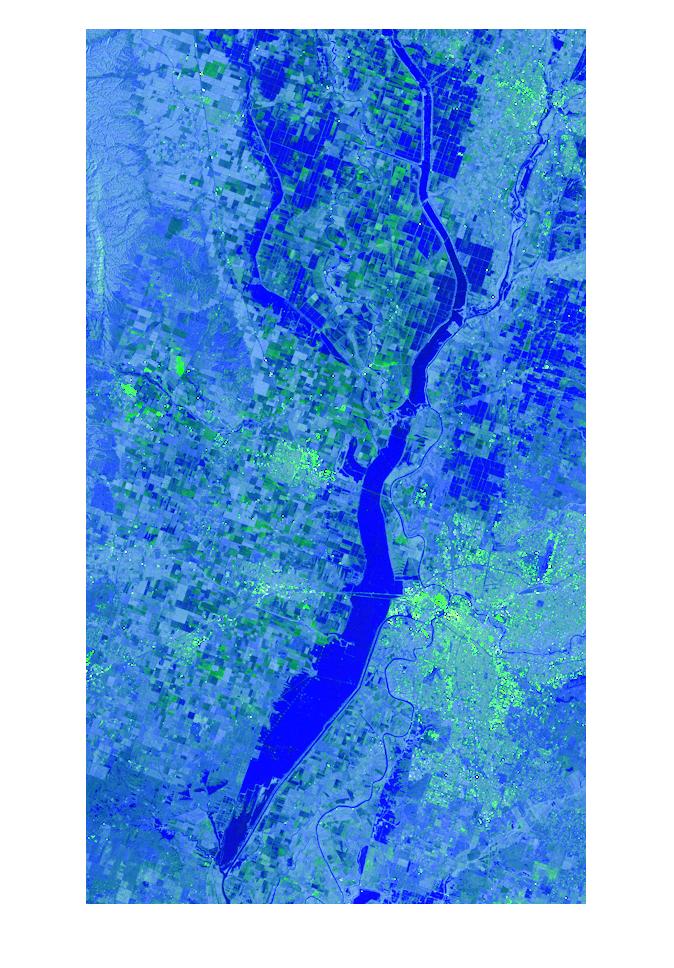}
\caption{Sentinel-1A ($t2$)}
\label{fig2:subim2}
\end{subfigure}
\begin{subfigure}[t]{0.24\textwidth}
\includegraphics[width=\linewidth,height=1.8\linewidth, trim = {6.3cm 4.9cm 6.3cm 2.3cm}, clip]{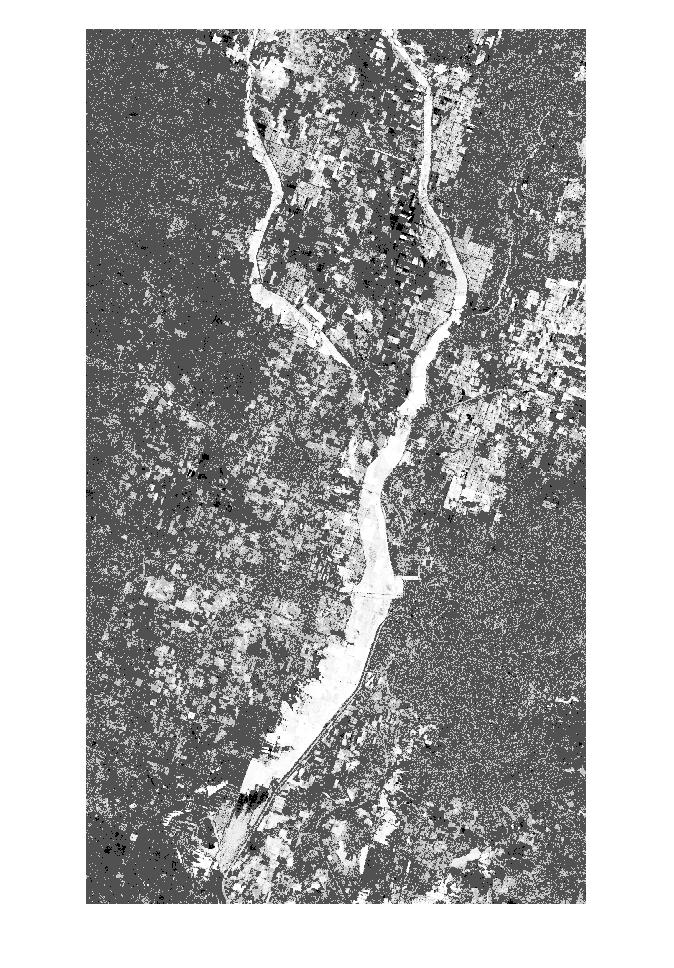}
\caption{Ratio $\nicefrac{SAR_{t_1}}{SAR_{t_2}}$}
\label{fig2:subim3}
\end{subfigure}
\begin{subfigure}[t]{0.24\textwidth}
\includegraphics[width=\linewidth,height=1.8\linewidth, trim = {6.3cm 4.9cm 6.3cm 2.3cm}, clip]{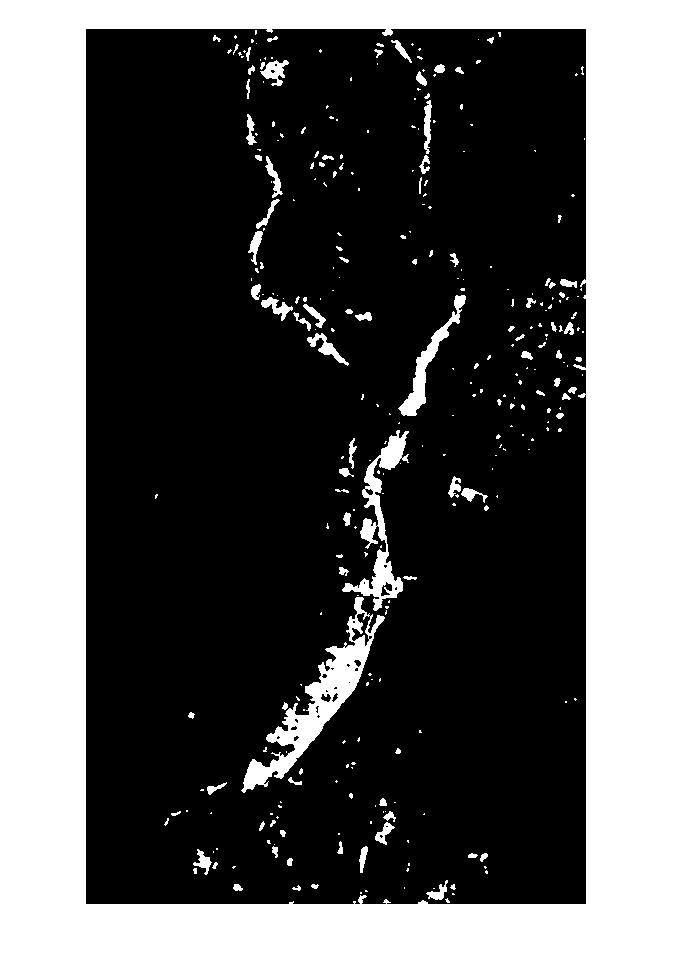}
\caption{Ground Truth}
\label{fig2:subim4}
\end{subfigure}

\caption{Flood in California: (a) Landsat 8 ($t1$), (b) Sentinel-1A ($t2$), (c) Ratio between SAR intensities at $t1$ and $t2$, (d) ground truth.}
\label{fig2:dataset2}
\end{figure*}

The second data set represents a more challenging scenario, as it involves an optical image and a synthetic aperture radar (SAR) image.
The image at time $1$ is a Landsat 8 acquisition covering Sutter County, California, on 5 January 2017\textsuperscript{1}.%\footnotemark[\ref{foot1}].
It is composed of $9$ channels covering the spectrum from deep blue to short-wave infrared, plus two long-wave infrared channels (Fig.\ \ref{fig2:subim1} shows the RGB channels).
Fig.\ \ref{fig2:subim2} shows the image acquired on 18 February 2017 by Sentinel-1A over the same scene, after the occurrence of a flood\footnote{Data processed by ESA, http://www.copernicus.eu/}.
The sensor uses two different polarisations (VV and VH), and the ratio between the two intensities completes the set of $3$ channels.
To obtain a reasonable ground truth without recourse to manual selection, we considered two other single-polarisation SAR images acquired approximately at the same times as the previous ones.
The normalised ratio between these images is depicted in Fig.\ \ref{fig2:subim3}, and the ground truth obtained by thresholding it at $0.5$ can be seen in Fig.\ \ref{fig2:subim4}.
The selection of an appropriate training set is not trivial.
There are many different kind of terrain involved, and excluding any of them might lead to poor results.
Therefore, the training set is drawn randomly from the parts of the images which are clearly distinguished as unchanged areas, to avoid to inadvertently exclude classes by a manual selection.
Again, the size of the set $\mathcal{T}$ is 2\% of the total image size.

The optimisation of the hyperparameters is carried out differently for every method.
Concerning the GP method, five iterations of the gradient ascent from random starting points are performed.
For the SVR method, a grid search on $\Omega_{\boldsymbol{\theta}}$ leads to selection of the best combination of $C$,$\epsilon$ and $\sigma$ after cross-validation.
A grid search for the RF hyperparameters investigates the values $T = \left\{32,64,128,256,512\right\}$ and $p = \left\{5,10,15,20\right\}$, while $m$ is kept equal to $\lfloor\nicefrac{P}{3}\rfloor$.
The same procedure is applied for the HPT over the values $K = \left\{16,32,64,128\right\}$ and $\gamma = \left\{10^{-2},\dots,10^{3}\right\}$.
A series of 100 runs is performed, each of these with a different training set, to evaluate the mean and standard deviation of the AUC and the training and test times for the four methods.
The results are summarised in Table \ref{tab:accuracy}.
\begin{table}[h!]
\centering
\caption{Mean and standard deviation of the AUC and the elapsed time for the four methods applied on the second data set.}
\label{tab:accuracy}
\begin{tabular}{ |c||c|c||c|c| }
\hline
\, & $m_{AUC}$ & $\sigma_{AUC}$ & $m_{t}$ & $\sigma_{t}$ \\
\hline
\hline
GP & $0.74692$ & $0.00043$ & $257.11$ & $1.47016$\\
\hline
RF  & $0.81680$ & $0.00541$ & $\boldsymbol{132.00}$ & $0.77075$\\
\hline
SVR  & $0.81299$ & $0.05455$ & $2024.58$ & $396.86244$\\
\hline
HPT  & $\boldsymbol{0.84001}$ & $0.01450$ & $924.91$ & $8.99086$\\
\hline
\end{tabular}
\end{table}

\begin{figure}[h!]
\centering
\includegraphics[width = \linewidth, keepaspectratio]{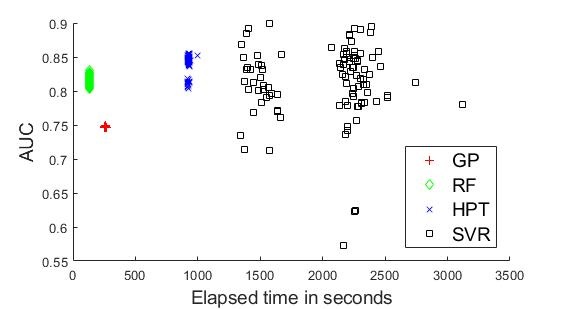}
\caption{Elapsed time vs AUC scatter plot for GP (red), SVR (black), RF (green), and HPT (blue)}
\label{fig:scatter_all}
\end{figure}

Similar results can be achieved by RF, SVR, and HPT.
On the contrary, the GP method produce worse results.
It could be thought that more iterations of the gradient ascent might lead to better solutions, but the consistency of the results throughout the whole series of runs suggests the opposite.
Instead, the main drawback of the SVR algorithm is the computing time.
Even a coarse grid search over the three hyper-parameters implicates a long training and validation phase.
Although it is capable of reaching peaks of $0.89$ for the AUC, it is also sensitive to the selection of the hyper-parameters.
This brings to a larger $\sigma_{AUC}$, and examples of low AUCs.
On the other hand, both the HPT and the RF method have their strong suit.
The former can reach better results, whereas the latter is computationally faster.
In Fig.\ \ref{fig:scatter_all}, the elapsed time vs AUC scatter plot for the series of runs supports the previous comments.

\begin{figure}[h!]
\centering
\includegraphics[width = \linewidth, keepaspectratio]{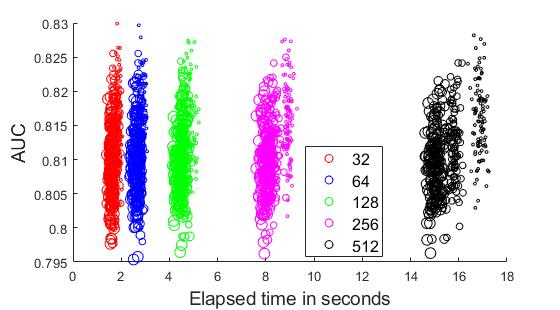}
\caption{Elapsed time vs AUC scatter plot for the RF: colors refer to different values of $T$ (see legend), whereas a bigger marker size denotes a larger number of considered features $m \in [5,10,15,20]$.}
\label{fig:scatter_rf}
\end{figure}

Focusing on the RF, the scatter plot in Fig.\ \ref{fig:scatter_rf} shows that the computational time grows linearly with respect to the number of trees.
However, a larger $T$ does not necessarily provide benefits, as this example demonstrates.
Instead, smaller values of $m$ brings on average to better results, at the cost of small additional times.
Consequently, it can be recommended to not exaggerate with the pruning.
We also recall that, in many formulations of the RF approach, trees are not pruned at all.
About the HPT method, no significant trends are recognised in its scatter plot (which is not reported), suggesting that the best option is to select a small number of neighbours $K$ to reduce the amount of computations, and to perform a log-scale search on $\gamma$ to find the most suitable.

\section{Conclusions}\label{sec:concl}

In this paper, we proposed a CD framework based on image regression and evaluated the performance obtained using four different regression methods.
The experiments on two data sets proved the effectiveness of the methodology, especially for two of the regression algorithms.
Although the HPT method achieved the best results, RF regression proved capable of reaching close results with a shorter computation time.
A future work would be to investigate further the role of the hyperparameters, also on other data sets, and to experiment with smaller or more difficult training sets.
Another important subject of future research will be the development of an unsupervised version of the methodology.

\bibliographystyle{IEEEbib}
\bibliography{references}

\end{document}